\documentclass{article}
\usepackage{amssymb}
\usepackage{times}

\usepackage{amsmath,amsfonts}
\usepackage{multicol}
\usepackage[bookmarks=true]{hyperref}
\usepackage{graphicx}
\usepackage{amsmath}
\usepackage{amsfonts} 
\usepackage{mathrsfs}
\usepackage{algorithm}      
\usepackage{algpseudocode}  
\usepackage[table]{xcolor} 
\usepackage{colortbl}
\usepackage{booktabs}
\usepackage{multirow}
\usepackage{latexsym}
\usepackage{caption}
\usepackage{comment}

\newcommand{\name}{LaDi-WM{}}

\newcommand{\ours}[1]{\textcolor{blue}{Method}~}

\usepackage[final]{corl_2025} 

\title{LaDi-WM: A Latent Diffusion-based World Model for Predictive Manipulation}

\author{
  Yuhang Huang\\
  National University of\\ 
  Defense Technology\\
  \And
  Jiazhao Zhang\\
  Peking University\\
  \And
  Shilong Zou\\
  National University of\\ 
  Defense Technology\\
  \And
  Xinwang Liu\\
  National University of\\ 
  Defense Technology\\
  \And
  Ruizhen Hu$^{*}$\\
  Shenzhen University\\
  \And
  Kai Xu\thanks{Co-corresponding author.}\\
  National University of\\ 
  Defense Technology\\
}

\begin{document}
\maketitle



    
\vspace{-20pt}
\begin{abstract}
    Predictive manipulation has recently gained considerable attention in the Embodied AI community due to its potential to improve robot policy performance by leveraging predicted states. However, generating accurate future visual states of robot-object interactions from world models remains a well-known challenge, particularly in achieving high-quality pixel-level representations.
    To this end, we propose LaDi-WM, a world model that predicts the latent space of future states using diffusion modeling. Specifically, LaDi-WM leverages the well-established latent space aligned with pre-trained Visual Foundation Models (VFMs), which comprises both geometric features (DINO-based) and semantic features (CLIP-based). We find that predicting the evolution of the latent space is easier to learn and more generalizable than directly predicting pixel-level images.
    Building on LaDi-WM, we design a diffusion policy that iteratively refines output actions by incorporating forecasted states, thereby generating more consistent and accurate results. Extensive experiments on both synthetic and real-world benchmarks demonstrate that LaDi-WM significantly enhances policy performance by 27.9\% on the LIBERO-LONG benchmark and 20\% on the real-world scenario. Furthermore, our world model and policies achieve impressive generalizability in real-world experiments. \textit{The source code will be public at: https://github.com/GuHuangAI/LaDiWM.}
\end{abstract}



\section{Introduction}



Predictive manipulation \citep{atm, avdc, unisim} has recently gained considerable attention in Embodied AI due to its potential to enhance robot policy performance by leveraging predicted future states. These predictive states, which represent anticipated interactions between robots and objects, have been widely received as beneficial for action planning which is error-prone especially for long-term tasks~\citep{dinowm, tdmpc2}. To obtain predictive states, most existing methods employ world models \citep{unisim, iris} to predict visual frames. However, pixel-level visual prediction often exhibits limited performance and generalizability. Another line of research advances world models by predicting latent representations \citep{dreamer, dreamerv2, dinowm}, which simplifies model learning and demonstrates strong generalizability. However, the latent space in existing works is typically optimized for image reconstruction, thus failing to capture the geometric and semantic information that is crucial for successful manipulations.

Inspired by latent-space world models, we propose \name, which is designed to efficiently model the correlation of geometric and semantic information in the latent space of Visual Foundation Models (VFMs). Our method incorporates both DINO-based~\citep{liu2024grounding, dinov2} and CLIP-based~\citep{goel2022cyclip, li2022blip, siglip} latent representations, capturing geometric and semantic information, respectively.
To effectively model their dynamic correlations, we introduce an interactive diffusion process that performs cross-attention between \emph{clean latent components} decomposed from noisy latent representations of geometry and semantics.
We find that modeling the dynamic interaction between geometry and semantics with latent space diffusion leads to better alignment between the two, and hence to fast learning convergence and strong cross-scene and cross-task generalibility.







\begin{figure*}
    \centering
    \includegraphics[width=\textwidth]{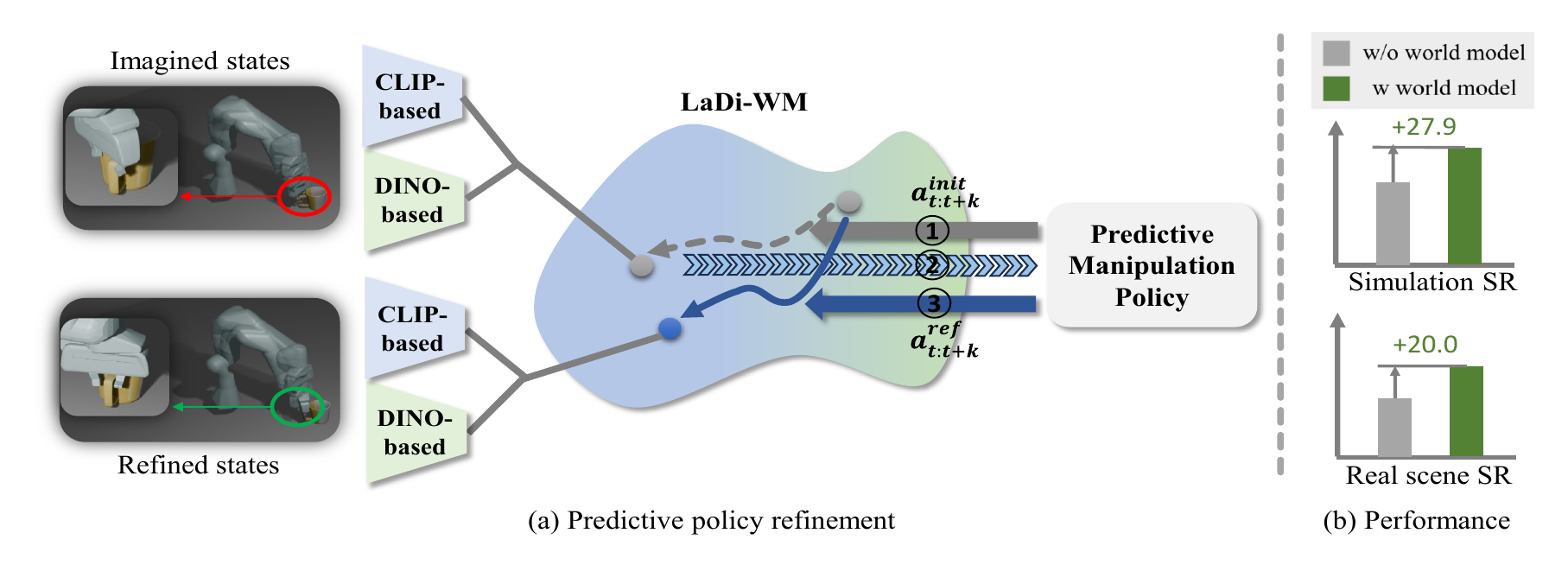}\vspace{-20pt}
    \vspace{-5pt}   
    \caption{Our method consists of a latent diffusion-based world model and a corresponding predictive manipulation policy. (a): \textcircled{1} Given an initial action sequence generated by the policy, our world model produces imagined future states. \textcircled{2} These imagined states provide efficient guidance for the policy model, \textcircled{3} yielding refined actions for improved manipulation. (b): Benefiting from our world model, we achieve significant improvements in both simulation and real-world performance.\vspace{-10pt}}
    \label{fig:teaser}
\end{figure*}
Based on \name, we propose an imagination-guided diffusion policy that features \emph{iterative action refinement}. As illustrated in Fig.~\ref{fig:teaser} (a), we train a diffusion policy within the latent space of both DINO and CLIP. This policy is trained to output a sequence of actions at a time. We predict for each action an imagined future state using the pretrained \name. These imagined states are then fed into the policy model to produce a refined sequence of actions. Such refinement is guided by multi-step imaginations, yielding higher quality actions. The refinement can be iterated for multiple times until convergence. We provide a discussion on the convergence of the iterative refinement and show that it progressively improves the accuracy and consistency of action predictions.



We have conducted extensive experiments on both synthetic benchmarks and real-world environments. Benefiting from the efficient future guidance by \name, our method achieves the highest success rate among the state-of-the-art methods. In particular, our approach significantly improves over the behavior cloning baseline by $27.9\%$ in the synthetic experiment and $20.0\%$ in the real-world environment, demonstrating the superior potential of LaDi-WM in guiding policy learning. 
\section{Related Work}

\textbf{World models} can be regarded as future prediction models that forecast future environmental states based on historical observations and action controls. 
Current studies on world models can be split into two types, which learn the dynamics in the raw image space and the latent space, respectively.
Typical world models \cite{ha2018world} employ recurrent neural networks \cite{schmidhuber1990line, schmidhuber1990reinforcement} and convolution networks \cite{racaniere2017imagination} to learn environmental dynamics as future image prediction. Advanced methods \cite{unisim, ding2024diffusion, alonso2024diffusion} explore the use of diffusion models to build world models and learn dynamics in pixel space; however, the computational expense associated with image reconstruction and the overhead of diffusion models limit their applicability.
In contrast, the Dreamer series of methods \cite{dreamer, dreamerv2, dreamerv3} builds upon recurrent neural networks and explores learning imagination in the latent space. Similarly, IRIS \cite{iris} and TDMPC2 \cite{tdmpc2} also learn latent dynamics but utilize transformer and MLP architectures for world modeling. 
 To enhance the representation ability, DINO-WM \cite{dinowm} leverages the pretrained DINO model to extract the latent representation containing rich geometric information. 
 In this paper, we propose the adoption of latent diffusion to establish a more generalizable world model, involving both geometric and semantic information, powered by pretrained foundation models.

\textbf{World models for predictive policy learning.} 
World modes can interact with the policy to generate future dynamics, which can be used to facilitate policy learning.
Early world models are usually utilized to support reinforcement learning. 
IRIS \cite{iris} and the Dreamer series \cite{dreamer, dreamerv2, dreamerv3}  propose learning policies in the imagined world to solve simple game tasks. TDMPC series methods \cite{tdmpc2} and DINO-WM \cite{dinowm} combine world models with model predictive control, optimizing action sequences based on a goal image. 
However, reinforcement learning \cite{schulman2017proximal, haarnoja2018soft} requires dense reward, limiting its application in relatively hard tasks such as manipulation.
Apart from reinforcement learning, another line of work lets the policy predict the action according to the future imagined observations.
AVDC \cite{avdc} employs a language-driven video prediction model to generate future frames and estimate future optical flow to calculate corresponding robotic transformations. Its performance depends largely on the accuracy of the estimated flow, which is not robust for manipulation tasks.
ATM \cite{atm} introduces an instruction-conditioned prediction model to predict future point trajectories and utilizes the prediction as the input to the policy network for behavior cloning. The task-specific prediction limits its generalization, and the point representation ignores the detailed information, which may constrain performance improvement.
In this paper, our world model is designed to deal with task-agnostic prediction for powerful generalization and learns more detailed information containing both geometry and semantics, resulting in a significant improvement in manipulation success rate.

\begin{figure*}
    \centering
    \includegraphics[width=\textwidth]{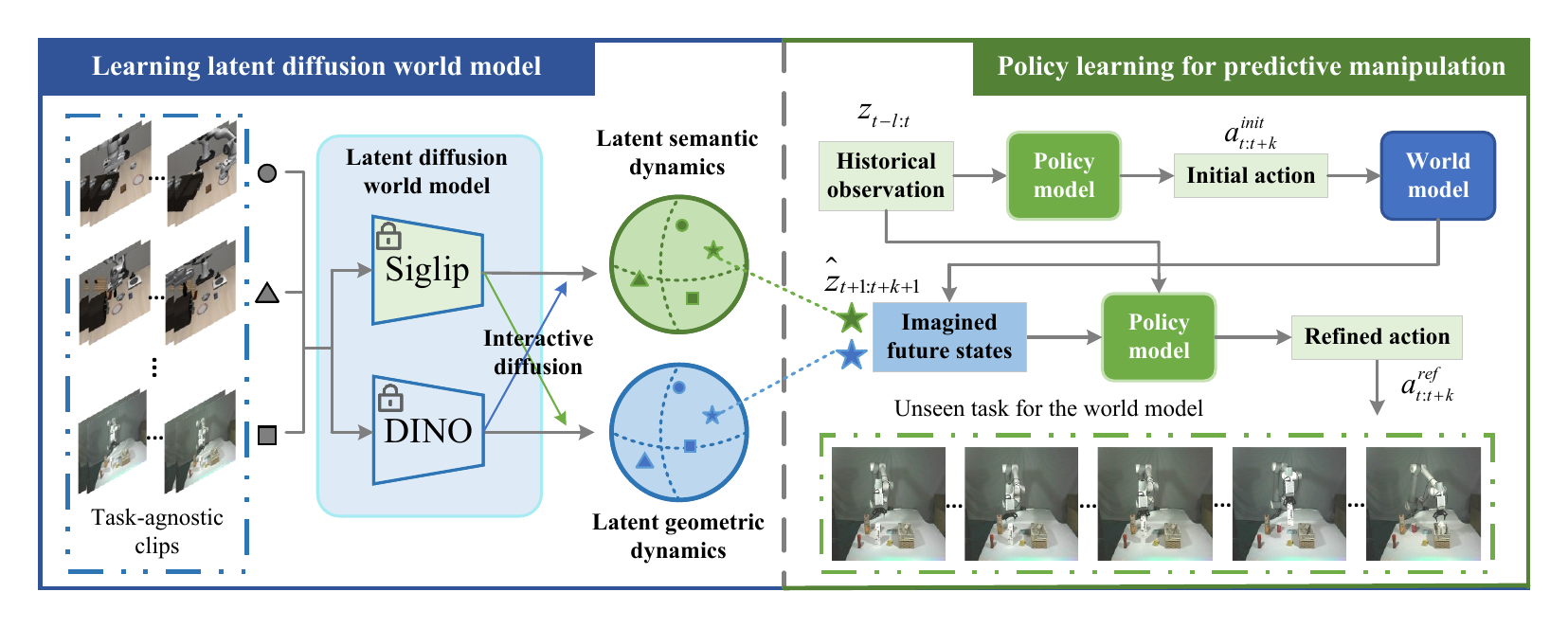}\vspace{-13pt}
    \caption{Overall framework of the proposed method. We first train LaDi-WM on task-agnostic clips, which benefits cross-task generalization. The trained world model is used to generate future imagined states, which serve as effective guidance and input to the policy network, improving the manipulation performance significantly.\vspace{-10pt} 
    }
    \label{fig:overview}
\end{figure*} 

\section{Method}
As illustrated in Fig.~\ref{fig:overview}, our method involves two components: a latent diffusion-based world model for imagining the future states, and an imagination-guided policy for action planning, with details provided in Sec.~\ref{sec:3.1} and Sec.~\ref{sec:3.2}, respectively. 

\subsection{Latent Diffusion-based World Model}\label{sec:3.1}
We propose to utilize the latent diffusion to construct the world model, aiming to capture the diverse generalizable dynamics of the environments.
Benefiting from the powerful ability of diffusion models for modeling multimodal distribution, our world model can handle complex environmental dynamics. 
Moreover, the diffusion process is conducted in a generalizable latent space that is governed by the pretrained foundation models, containing both geometric and semantic information.

Specifically, we adopt DINO \cite{dinov2} and Siglip \cite{siglip} to extract both geometric and semantic latent codes, which are crucial for executing robotic tasks. The latent states extracted via the foundation models can be expressed by:
\begin{equation}
    \mathbf{z}_{t} = [{\mathbf{z}^{D}_{t}}; {\mathbf{z}^{S}_{t}}] = [f_{dino}(I_{t});f_{sigl}(I_t)],
\end{equation}
where $f_{dino/sigl}$ represents the DINO and Siglip networks, $I_t$ and $\mathbf{z}_{t}$ are the image observation and latent state at time $t$.


After obtaining the latent states $\mathbf{z}_{t}$, the goal of the world model is to predict future latent states given historical latent states $\mathbf{z}_{t-l:t}$ and action control $a_{t:t+k}$: 
\begin{equation}
        \mathbf{z}_{t+1:t+k+1} \sim p_{\boldsymbol{\theta}}(\mathbf{z}_{t+1:t+k+1}|\mathbf{z}_{t-l:t}, a_{t:t+k}),
\end{equation}
where $p$ is the transition probability distribution, and $\boldsymbol{\theta}$ is the parameter of the world model.
To simplify the following formulation, we replace the future latent states $\mathbf{z}_{t+1:t+k+1}$ with $\mathbf{z}_{t+1:}$.


\textbf{Interactive diffusion modeling}. We propose to utilize the diffusion model to model the transition probability. However, the two types of codes follow different distributions, 
which implies that it is inappropriate to apply the one diffusion process to learn the two types of dynamics. 
On the other hand, a separate learning framework may overlook the relationships between high-level semantics and low-level geometry, resulting in inefficient dynamic modeling. 
To this end, we propose a novel interactive diffusion process that enables two latent codes to interact with each other, leading to more effective dynamic modeling. 
Moreover, inspired by \citet{ddm}, we decompose the clean latent component from the noisy codes and leverage the clean latent component to conduct interaction. 


In more detail, the forward diffusing process consists of an attenuation process of the latent states and a noise-increasing process. To formulate the forward process, we introduce an additional superscript $n$ to represent the diffusion step, i.e., we use ${\mathbf{z}_{}}_{t+1:,0}= \mathbf{z}_{t+1:}$ and $\mathbf{z}_{t+1:,n}$ to denote the clean latent states and noisy states.
Considering a continuous-time diffusion equation with $n\in[0, 1]$, the forward process for the DINO latent code can be represented by:
\begin{equation}
\begin{aligned}
    {\mathbf{z}^{D}_{t+1:, n}} = {\mathbf{z}^{D}_{t+1:,0}} + \int_0^n \boldsymbol{C}^{D}_{n} \mathrm{d}n + \sqrt{n}\boldsymbol{\eta}^{D}, \boldsymbol{\eta}^{D}\sim\mathcal{N}(\mathbf{0}, \mathbf{I}),
\end{aligned}
\end{equation}
where ${\mathbf{z}^{D}_{t+1:,0}} + \int_0^n \boldsymbol{C}^{D}_{n} \mathrm{d}n$ represents the attenuation process of the latent states while $\sqrt{n}\boldsymbol{\eta}^D$ denotes the growth of the normal gaussian noise, and ${\mathbf{z}^{D}_{t+1:,n}}$ is the noisy latent code. 
Essentially, ${\boldsymbol{C}^{D}_{n}}$ is the gradient of the latent code attenuation, which can be solved by ${\mathbf{z}^{D}_{t+1:,n}} + \int_0^1 \boldsymbol{C}^{D}_{n} \mathrm{d}n = \boldsymbol{0}$. 
According to the derivation in \cite{ddm}, the attenuation process is governed by $\boldsymbol{C}^{D}_{n}$, thus we can use $\boldsymbol{C}^{D}_{n}$ as the clean component of $\mathbf{z}^{D}_{t+1:,n}$. 
Similarly, we can denote the clean component of $\mathbf{z}^{S}_{t+1:,n}$ as $\boldsymbol{C}^{S}_{n}$.


Following the derivation in \citet{ddm}, the conditional probability used in the reversed process can be expressed by:
\begin{equation}
\begin{aligned}
    q({\mathbf{z}^{D}_{t+1:,n-\Delta n}}&|{\mathbf{z}^{D}_{t+1:,n}}, {\mathbf{z}^{D}_{t+1:,0}}) = \mathcal{N}({\mathbf{z}^{D}_{t+1:,n-\Delta n}}, \boldsymbol{\mu}, \boldsymbol{\Sigma}),\\
    &\boldsymbol{\mu} = {\mathbf{z}^{D}_{t+1:,n}}+\Delta t{\mathbf{z}^{D}_{t+1:,0}}-\frac{\Delta n}{\sqrt{n}}\boldsymbol{\eta}^{D},\\
    &\boldsymbol{\Sigma} = \frac{\Delta n(n-\Delta n)}{n}\mathbf{I}.
\end{aligned}
\end{equation}
In the reversed process, the clean latent code ${\mathbf{z}^{D}_{t+1:,0}}$ and $\boldsymbol{\eta}$ are not accessible, hence we need to use the parameterized $p_{\boldsymbol{\theta}}({\mathbf{z}^{D}_{t+1:,n-\Delta n}}|{\mathbf{z}^{D}_{t+1:,n}})$ to approximate $q({\mathbf{z}^{D}_{t+1:,n-\Delta n}}|{\mathbf{z}^{D}_{t+1:,n}}, {\mathbf{z}^{D}_{t+1:,0}})$.
In this way, the denoising network outputs ${\mathbf{z}^{D}_{t+1:,\boldsymbol{\theta}}}$ and $\boldsymbol{\eta}^{D}_{\boldsymbol{\theta}}$ to parameterize ${\mathbf{z}^{D}_{t+1:,0}}$ and $\boldsymbol{\eta}^{D}$ simultaneously. Note that we perform a similar implementation to predict the future Siglip latent code ${\mathbf{z}^{S}_{t+1:,0}}$.

To let the two types of latent codes interact with each other in the diffusion process, we first leverage the decomposition network $f_{\boldsymbol{\theta}_1}$ and $f_{\boldsymbol{\theta}_2}$ to estimate the clean components $\boldsymbol{C}^{D}_{n}$ and $\boldsymbol{C}^{S}_{n}$. The outputs can be expressed via:
\begin{equation}
    \begin{aligned}
        &\boldsymbol{C}^{D}_{\boldsymbol{\theta}} = f_{\boldsymbol{\theta}_1}({\mathbf{z}^{D}_{t+1:,n}}, n, {\mathbf{z}^{D}_{t-l:t}}, a_{t:t+k}),\\
        &\boldsymbol{C}^{S}_{\boldsymbol{\theta}} = f_{\boldsymbol{\theta}_2}({\mathbf{z}^{S}_{t+1:,n}}, n, {\mathbf{z}^{S}_{t-l:t}}, a_{t:t+k}).
    \end{aligned}
\end{equation}
Here, ${\mathbf{z}^{D}_{t-l:t}}$ and ${\mathbf{z}^{S}_{t-l:t}}$ are the historical latent codes and $a_{t:t+k}$ is the given action control, which serves as the conditional inputs.
Then, $\boldsymbol{C}^{D}_{\boldsymbol{\theta}_1}$ and $\boldsymbol{C}^{S}_{\boldsymbol{\theta}_2}$ are fed into the denoising network $f_{\boldsymbol{\theta}_3}$ and $f_{\boldsymbol{\theta}_4}$, generating the parameterized ${\mathbf{z}^{D}_{t+1:, \boldsymbol{\theta}}}$ and $\boldsymbol{\eta}^{D}$ for recovering future latent codes.
\begin{equation}
\begin{aligned}
    &{\mathbf{z}^{D}_{t+1:,\boldsymbol{\theta}}}, \boldsymbol{\eta}^{D}_{\boldsymbol{\theta}} = f_{\boldsymbol{\theta}_3}({\mathbf{z}^{D}_{t+1:,n}}, n, \boldsymbol{C}^{S}_{\boldsymbol{\theta}}, a_{t:t+k}),\\
    &{\mathbf{z}^{S}_{t+1:,\boldsymbol{\theta}}}, \boldsymbol{\eta}^{S}_{\boldsymbol{\theta}} = f_{\boldsymbol{\theta}_4}({\mathbf{z}^{S}_{t+1:,n}}, n, \boldsymbol{C}^{D}_{\boldsymbol{\theta}}, a_{t:t+k}).
\end{aligned}
\end{equation}


We adopt the diffusion transformer \citep{peebles2023scalable} architecture to construct our world model, and more details can be found in Appendix~\ref{apd:1}.




\subsection{Predictive Policy Learning}\label{sec:3.2}
Based on the world model, we propose a predictive policy learning framework to efficiently solve manipulation tasks by imitation learning. 
Specifically, given an initial action generated by the policy model, the world model can imagine the future latent states based on historical states.
The imagined future states indicate the possible outcomes caused by the initial action, which provides effective guidance for the policy model. Utilizing the imagined guidance as additional input, the policy model can predict the action precisely and refine itself iteratively during testing time:
\begin{equation}
\label{eq2}
    a_{t:t+k} = \pi_{\boldsymbol{\theta}}(\mathbf{z}_{t-l:t}, \mathbf{e}_{t-l:t}, \hat{\mathbf{z}}_{t+1:t+k+1}, \mathcal{T}),
\end{equation}
where $\mathbf{z}_{t-l:t}$ and $\mathbf{e}_{t-l:t}$ are $l$-frame historical latent and end-effector states, $\hat{\mathbf{z}}_{t+1:t+k+1}$ is the imagined $k$-frame future states, and $\mathcal{T}$ is the task instruction.


We employ a diffusion policy with the transformer architecture as our policy model. The diffusion policy has advantages in learning multimodal action distributions and addressing high-dimensional action spaces, which is beneficial for solving diverse manipulation tasks.
We build the diffusion process following \citet{ddm} to improve the efficiency of action prediction.
We utilized the different modalities of historical observations as input, including task instruction, diffusion time step, robotic end-effector states, and historical and imagined latent states. 

Concretely, given the task instruction $\mathcal{T}$, the diffusion time step $n$, the historical $l$ frames of latent representation $\mathbf{z}_{t-l:t}$ and end-effector poses $\mathbf{e}_{t-l:t}$, we first utilize the policy model to predict an initial action sequence $\hat{a}_{t:t+k-1}$. The inital action sequence is fed into the world model with $\mathbf{z}_{t-l:t}$, producing the imagined future states $\hat{\mathbf{z}}_{t+1:t+k}$. Then, we utilize $\hat{\mathbf{z}}_{t+1:t+k}$ as additional inputs of the policy model, guiding the prediction of the denoising process. More specifically, we tokenize different modalities of inputs and concatenate them together, and utilize the transformer encoder to fuse the different modality information. The fused features are sent to the transformer decoder, accompanied by the noisy action, obtaining the denoised action. 
The detailed architecture of the policy model can be found in Appendix~\ref{apd:1}.

\textbf{Iterative refinement in inference time}.
During inference, our policy model allows for an iterative refinement process through multi-time exploration within the world model.
When an initial action is generated by the policy model, the world model produces corresponding future state predictions. However, since the initial action may be suboptimal, the resulting imagined future states may lack sufficient precision to provide effective guidance. To address this limitation, we implement an iterative correction mechanism in which the policy model continuously refines its actions based on progressively more accurate imagined states generated by the world model. This closed-loop interaction between policy refinement and world model prediction enables increasingly precise action optimization, ultimately leading to more robust and effective policy execution. The detailed inference procedure is provided in Appendix~\ref{apd:3}. We also analyze the convergence of the iterative refinement, which can be found in Appendix~\ref{ape:convergence}. 
\section{Experiments}
We conduct simulated experiments on two public benchmarks, LIBERO-LONG \citep{libero} and CALVIN D-D \citep{mees2022calvin}.
The detailed information about the two benchmarks and evaluation metrics can be found in Appendix~\ref{apd:2}.
The quantitative comparisons to baselines, 
ablation studies, and real-world experiments are provided in Sec.~\ref{sec:main_result},  Sec.~\ref{sec:ablation_study}, and Sec.~\ref{sec:real_exp}, respectively.

\begin{table}[t]
    \centering
    \caption{\textbf{Performance on LIBERO-LONG benchmark.}}
    \label{tab:1}
    \renewcommand{\arraystretch}{1.2}
    \setlength{\tabcolsep}{3pt}
    \resizebox{1\linewidth}{!}{
    \begin{tabular}{l|cccccccccc|c}
    \toprule
    Method  & Task1 & Task2 & Task3 & Task4 & Task5 &Task6 & Task7& Task8& Task9&Task10 & Avg.SR$\uparrow$ \\\hline\hline
    DreamerV3 \citep{dreamerv3}  & 38.3 & 21.7 & 33.3 & 26.7  & 30.0 & 28.3  & 41.7 &  31.7 & 33.3 & 50.0 & 33.5  \\
    TDMPC2  \citep{tdmpc2}  & 45.0 & 36.7 & 35.0  & 31.7 & 31.7 &  23.3  &  36.7 & 40.0 & 38.3 & 51.7 & 37.0   \\
    ATM  \citep{atm}    & 46.7 & 58.3  &  60.0  &  31.7  & 33.3 & 20.0 & 43.3  & 51.7 & 41.7  & 53.3 & 44.0  \\
    Seer  \citep{seer}   & 71.7  &  50.0  &  48.3  & \textbf{51.7} & 66.7 & 53.3 & 51.7 & 45.0 & 48.3 & 50.0 &  53.6\\
    \rowcolor{gray!20}
    Ours   & \textbf{88.3} & \textbf{68.3}  & \textbf{63.3} & 45.0  & \textbf{83.3} & \textbf{65.0}  & \textbf{78.3}  &  \textbf{63.3}   &  \textbf{60.0}  &  \textbf{71.7}  & \textbf{68.7} \\     
    \bottomrule
    \end{tabular}
    }
    \vspace{-10pt}
\end{table}

\subsection{Quantitative Results}
\label{sec:main_result}



\textbf{LIBERO-LONG benchmark results.} 
The results on the Libero benchmark are shown in Tab.~\ref{tab:1}, where we report detailed performance for each task along with the average success rate, with all models trained using only 10 demonstrations per task.
Notably, our proposed world model achieves significant improvements in policy learning, despite being trained on dynamics unseen during policy training. Compared to the previous SOTA~\cite{seer}, our approach improves the average success rate by $15.1\%$. This advancement demonstrates the effectiveness of our framework. We provide more results of LIBERO-LONG benchmark in Appendix~\ref{apd:full_compare}.

\textbf{CALVIN D-D benchmark results.} 
Tab.~\ref{tab:calvin} reports the performance comparison on the CALVIN D-D benchmark. Firstly, compared to vanilla behavior cloning (BC) without the world model, our method significantly improves the Avg.Len. by 1.19. Moreover, our method surpasses other methods in terms of Avg.Len. and success rate, demonstrating the superiority of the predictive policy learning.


\begin{table*}[t]
    \centering
    \caption{\textbf{Performance on CALVIN D-D benchmark.}}
    \label{tab:calvin}
    \renewcommand{\arraystretch}{1}
    \setlength{\tabcolsep}{8pt}
\begin{tabular}{l|cccccc}
\toprule
\multirow{2}{*}{Method} & \multicolumn{6}{c}{Task completed in a row}       \\ \cline{2-7} 
                        & 1 & 2 & 3 & 4 & \multicolumn{1}{c|}{5} & Avg.Len.$\uparrow$ \\ \hline\hline
Vanilla BC                     & 81.4  & 60.8  & 45.9  & 32.1  & \multicolumn{1}{c|}{24.2}  &  2.44    \\
DreamerV3  \citep{dreamerv3}                    & 82.0  & 63.1  & 46.6  & 34.1  & \multicolumn{1}{c|}{25.1}  &  2.51    \\
ATM   \citep{atm}                   & 83.3  & 70.9  & 58.5  & 46.3  & \multicolumn{1}{c|}{39.1}  &  2.98    \\
Seer  \citep{seer}                   & 92.2  & 82.6 &  71.5 & 60.6  & \multicolumn{1}{c|}{53.5}  &  3.60    \\
\rowcolor{gray!20}
\rowcolor{gray!20}
Ours                    & \textbf{92.7} & \textbf{83.1}  &  \textbf{72.1} & \textbf{61.2}  & \multicolumn{1}{c|}{\textbf{54.1}}  &  \textbf{3.63}    \\ 
\bottomrule
\end{tabular}\vspace{-15pt}
\end{table*}

\begin{figure}[tb]
    \centering
    \includegraphics[width=\linewidth]{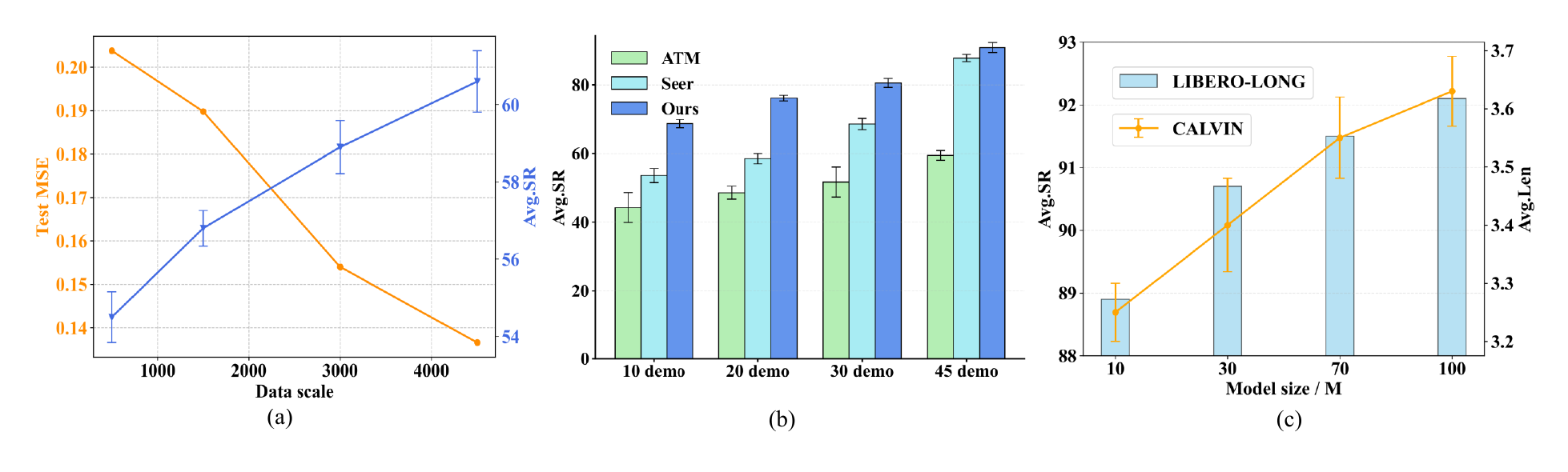}
    \vspace{-25pt}\caption{(a) scaling up the training data of the world model; (b) scaling up the training data of the policy model; (c) scaling up the model size of the policy model.}
    \label{fig:data-scale}
    \vspace{-10pt}
\end{figure}

\textbf{Scalability}.
Figure~\ref{fig:data-scale} demonstrates our method's scalability across three dimensions. 
For world model training, we evaluate performance using 500 to 4,500 demonstrations. Our results show that increasing training data progressively reduces test MSE while simultaneously improving policy learning performance.
In policy learning, expanding the training set from 10 demonstrations to complete data yields corresponding improvements in success rate. Notably, our approach consistently outperforms the previous SOTA~\cite{seer} across all data scales, ultimately achieving a 90.7\% average success rate (Avg.SR), i.e., a 3\% improvement over Seer's best performance.
Furthermore, scaling up the policy model architecture produces measurable gains on both the LIBERO-LONG and CALVIN benchmarks, further validating our method's scalability.


\subsection{Ablation studies}\label{sec:ablation_study}
We conducted a series of ablation experiments in simulated environments to evaluate the impact of our design choices.
Except for necessary explanations, the ablations are conducted on the LIBERO-LONG with 10 training demonstrations for each task. 




\textbf{Ablation study of world model architecture}. As shown in the first row of Tab.~\ref{tab:4}, our method outperforms the transformer-only architecture by 7.8\%.  As shown in the second row of Tab.~\ref{tab:4}, incorporating semantic information leads to a 3.4\% improvement in average success rate, particularly for Task 1, Task 5, and Task 9, which involve objects with distinct semantic features. As shown in the third row, the interactive diffusion process outperforms the concatenation method by 1.8\%, demonstrating its ability to learn more accurate dynamic predictions.

\begin{table}[!t]
    \centering
    \caption{\textbf{Ablation study of different world model architecture}, evaluated with 1-time refinement.}
    \label{tab:4}
    \renewcommand{\arraystretch}{1.5}
    \setlength{\tabcolsep}{4pt}
    \resizebox{1\linewidth}{!}{
    \begin{tabular}{l|cccccccccc|c}
    \toprule        
    Method  & Task1 & Task2 & Task3 & Task4 & Task5  &Task6 & Task7& Task8& Task9&Task10 & Avg.SR \\\hline\hline
    without diffusion    & 78.3 &  75.0 &  45.0  &  40.0   & 51.7 & 50.0 & 45.0  & 66.7 & 35.0 & 35.0 & 52.1 \\
    without siglip   & 85.0 &  81.7 &  \textbf{51.7}  &  45.0   & 60.0 & 50.0 & 51.7  & 68.3 & 40.0 & 40.0 & 57.3  \\
    without interaction   &  \textbf{90.0} &  80.0 &  46.7  &  48.3   & 61.7 & 51.7 & \textbf{55.0}  & 70.0 & 40.0 & \textbf{46.7} & 58.9 \\
    without clean interaction   &  \textbf{83.3} &  80.0 &  43.3  &  45.0   & 60.0 & 53.3 & 50.0  & 70.0 & 41.7 & 40.0 & 56.7 \\
    \rowcolor{gray!20}
    Ours   & \textbf{90.0} &  \textbf{83.3} &  50.0  &  45.0   & \textbf{66.7} & \textbf{55.0}  & \textbf{55.0}  & \textbf{71.7} & \textbf{50.0} & 40.0 & \textbf{60.7} \\\bottomrule
    \end{tabular}
    }
\end{table}

\begin{figure}
    \centering\vspace{-15pt}
    \includegraphics[width=\linewidth]{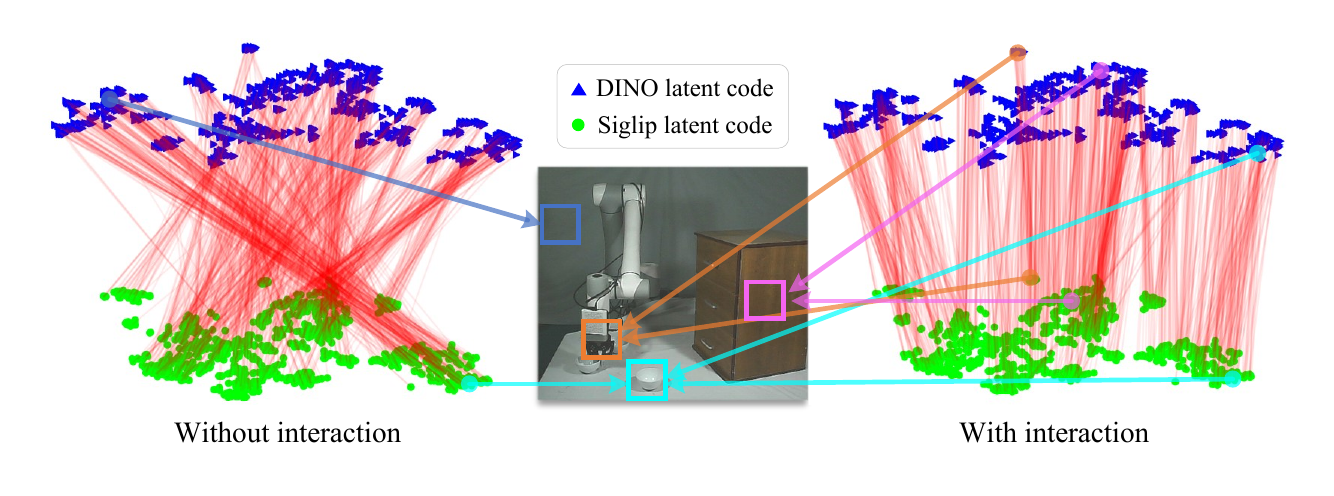}
    \vspace{-25pt}
     \caption{Visual analysis of interactive diffusion. Without the interactive diffusion, the latent codes exhibit connections to distinct regions (left). In contrast, interactive diffusion consistently aligns connections to identical objects (right).\vspace{-20pt}}
    \label{fig:inter_att}
\end{figure}

We plot the learned relationship between the two types of dynamics in Fig.~\ref{fig:inter_att}. According to the attention weight of the transformer layer, the most related features are connected by red lines. The result indicates that our interactive diffusion facilitates the alignment of two latent distributions.

\begin{table}[t]
    \centering
    \caption{\textbf{Ablation study of iteration in refinement.}}
    \label{tab:refine}
    \renewcommand{\arraystretch}{1.2}
    \setlength{\tabcolsep}{3pt}
    \resizebox{1\linewidth}{!}{
    \begin{tabular}{l|cccccccccc|c}
    \toprule
    Method  & Task1 & Task2 & Task3 & Task4 & Task5 &Task6 & Task7& Task8& Task9&Task10 & Avg.SR$\uparrow$ \\\hline\hline
    without imagination    &  66.7 &  63.3 &  31.7  &  40.0   & 35.0 & 41.7 & 35.0  & 43.3 & 25.0 & 26.7 & 40.8  \\
    1-iter imagination  &  \textbf{90.0} &  \textbf{83.3} &  50.0  &  45.0   & 66.7 & 55.0 & 55.0  & \textbf{71.7} & 50.0 & 40.0 & 60.7 (+19.9) \\
    2-iter imagination &  88.3 &  68.3 &  \textbf{63.3}  &  45.0   & \textbf{83.3} & \textbf{65.0} & \textbf{78.3}  & 63.3 & \textbf{60.0} & \textbf{71.7} & \textbf{68.7 (+27.9)}  \\
    3-iter imagination &  \textbf{90.0} &  76.7 &  58.3  &  \textbf{48.3}   & 78.3 & 63.3 & 71.7  & 66.7 & 58.3 & 68.3 & 67.9 (+27.1) \\
    4-iter imagination &  \textbf{90.0} &  81.7 &  60.0  &  46.7   & 75.0 & 61.7 & 71.7  & 70.0 & 58.3 & 66.7 & 68.2 (+27.4)\\
    \bottomrule
    \end{tabular}
    }
\end{table}

\textbf{Iterative refinement}.
We conduct an ablation study on the LIBERO-LONG benchmark to investigate the impact of interaction frequency during iterative refinement. The results are presented in Tab.~\ref{tab:refine}, revealing that increasing the number of iterations (from 0 to 2) yields significant performance gains ($27.9\%$ in Avg. SR). However, performance plateaus beyond 2 iterations (3 to 4), suggesting that two rounds of imagination suffice for our model.

\textbf{Other experiments}. We also ablate the effects of latent diffusion 
and the influence of different imagined future frames. More details can be found in Appendix~\ref{apd:ablation}.

\subsection{Real-world Experiments}\label{sec:real_exp}
\begin{figure*}[!t]
    \centering\vspace{-10pt}
    \includegraphics[width=1.\textwidth]{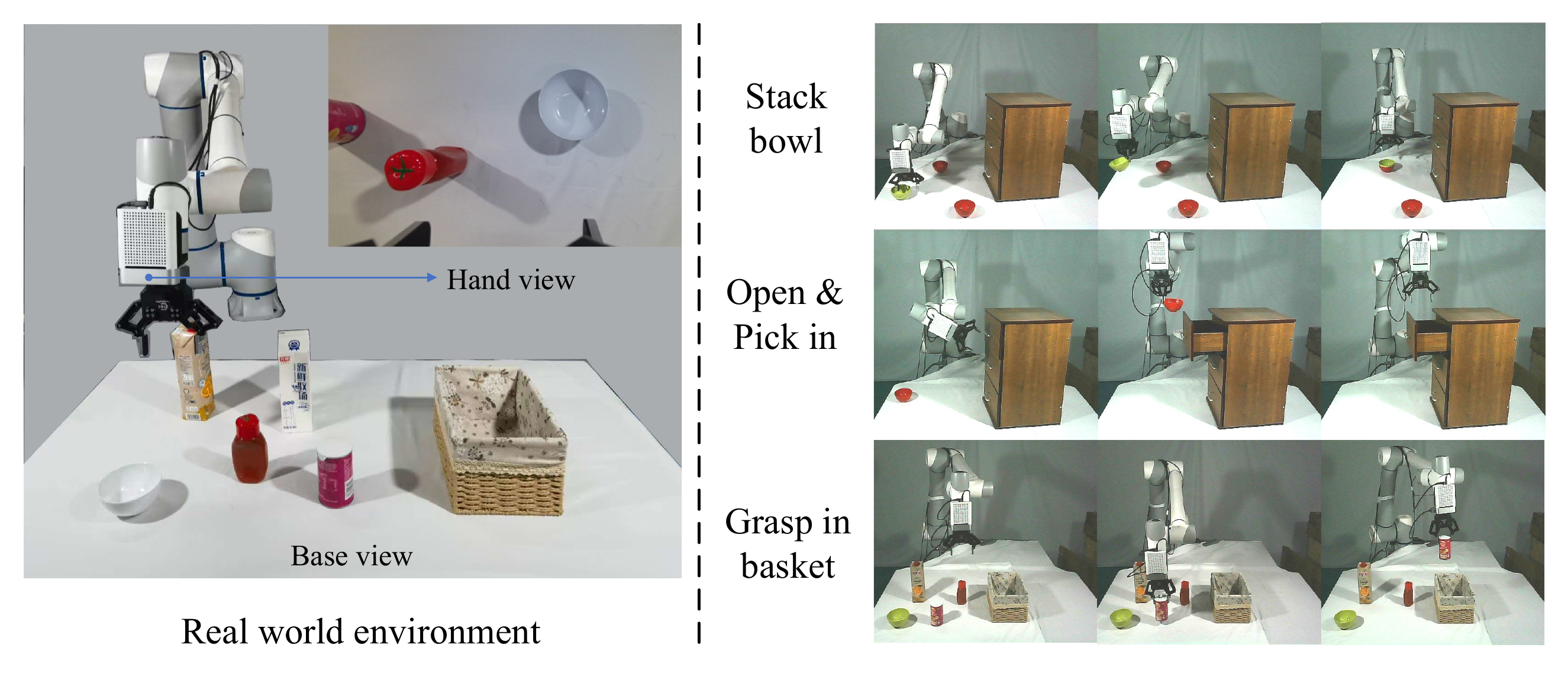}
    \vspace{-15pt}
    \caption{Real-world experiments: left shows the setup and right shows several execution examples.\vspace{-10pt}}
    \label{fig:real-setup}
\end{figure*}

We conduct the manipulation experiment in the real world with a 7-DOF CR10 robot arm, as shown in Fig.~\ref{fig:real-setup}. Same as in the simulated environment, we adopt two cameras to collect images from two views. The action space is the delta of the OSC pose. 
We define 7 tasks (details can be found in Appendix~\ref{apd:real}) and use the same metric as LIBERO-LONG to evaluate our method.
We finetune the policy model that is trained on the simulated environment using 10 realistic demonstrations per task. 
\begin{table}[!t]
    \centering
    \caption{\textbf{Real-world performance of variants.}}
    \label{tab:real}
    \renewcommand{\arraystretch}{1}
    \setlength{\tabcolsep}{5pt}
    \begin{tabular}{l|ccccccc|c}
    \toprule        
    Method  & Task1 & Task2 & Task3 & Task4 & Task5  &Task6 & Task7 & Avg.SR \\\hline\hline
    Vanilla BC    & 35.0 &  25.0 &  15.0  &  45.0   & 55.0 & 50.0 & 55.0  & 40.0 \\
    \rowcolor{gray!20}
    Ours   & \textbf{55.0} &  \textbf{45.0} &  \textbf{35.0}  &  \textbf{65.0}   & \textbf{75.0} & \textbf{70.0} & \textbf{75.0}  & \textbf{60.0}  \\
    \rowcolor{gray!20}
    Ours (pixel diffusion)  & 40.0 &  30.0 &  25.0  &  \textbf{65.0}   & 65.0 & 65.0 & 70.0  &  51.4 \\
    \rowcolor{gray!20}
    Ours (without diffusion)  & 35.0  & 25.0 &  25.0 &  60.0   & 65.0 & 65.0 & 70.0  & 49.3 \\\bottomrule
    \end{tabular}
\end{table}

    \vspace{-10pt}

Tab.~\ref{tab:real} reports the detailed performance comparison. Compared to the vanilla behavior cloning method, our policy model obtains efficient guidance via imagination of the world model, achieving an improvement of 20\% in average success rate. Moreover, we also ablate the latent and the diffusion designs in real-world experiments. The latent design and the diffusion design enhance the Avg.SR by 8.6\% and 10.7\% respectively.

\vspace{-5pt}
\section{Conclusion} 

We introduce LaDi-WM, a novel Latent Diffusion World Model that harnesses pre-trained visual foundation models to predict robotic future states in latent space. Our framework establishes distinct latent representations through DINO (geometric features) and SigLip (semantic features), combined with an iterative diffusion process that effectively models complex dynamics to generate high-fidelity future state predictions. The tight integration between LaDi-WM and diffusion policies enables enhanced policy learning through latent-space future state conditioning. Comprehensive evaluations across both synthetic and real-world benchmarks demonstrate LaDi-WM's superior performance in augmenting diffusion-based policy effectiveness. These results demonstrate the significant potential of leveraging established latent representations for developing robust world models, advancing toward robotic systems capable of reliable long-horizon prediction. 

\newpage
\section{Limitations}
We propose a latent diffusion world model to learn general environmental dynamics, which can provide future imaginations as efficient guidance for solving various manipulation tasks. One limitation of our method lies in the limited scale of training samples. We plan to incorporate more training data~\cite{o2024open, khazatsky2024droid} from diverse environments and tasks. On the other hand, longer-term future predictions can intuitively provide superior guidance for policy learning of long-horizon tasks, yet current world models face fundamental limitations in generating accurate extended predictions due to compounding error accumulation. To address this challenge, we identify the integration of memory mechanisms for maintaining long-horizon temporal consistency as a promising direction for future research.

\section{Acknowledgement}
This work is supported in part by the NSFC (62325211, 62132021, 62372457, 62322207), the Major Program of Xiangjiang Laboratory (23XJ01009).



\bibliography{references}  

\begin{thebibliography}{31}
\providecommand{\natexlab}[1]{#1}
\providecommand{\url}[1]{\texttt{#1}}
\expandafter\ifx\csname urlstyle\endcsname\relax
  \providecommand{\doi}[1]{doi: #1}\else
  \providecommand{\doi}{doi: \begingroup \urlstyle{rm}\Url}\fi

\bibitem[Wen et~al.(2024)Wen, Lin, So, Chen, Dou, Gao, and Abbeel]{atm}
C.~Wen, X.~Lin, J.~I.~R. So, K.~Chen, Q.~Dou, Y.~Gao, and P.~Abbeel.
\newblock Any-point trajectory modeling for policy learning.
\newblock In \emph{Robotics: Science and Systems}, 2024.

\bibitem[Ko et~al.(2024)Ko, Mao, Du, Sun, and Tenenbaum]{avdc}
P.~Ko, J.~Mao, Y.~Du, S.~Sun, and J.~B. Tenenbaum.
\newblock Learning to act from actionless videos through dense correspondences.
\newblock In \emph{International Conference on Learning Representations, {ICLR}}, 2024.

\bibitem[Yang et~al.(2024)Yang, Du, Ghasemipour, Tompson, Kaelbling, Schuurmans, and Abbeel]{unisim}
S.~Yang, Y.~Du, S.~K.~S. Ghasemipour, J.~Tompson, L.~P. Kaelbling, D.~Schuurmans, and P.~Abbeel.
\newblock Learning interactive real-world simulators.
\newblock In \emph{International Conference on Learning Representations, {ICLR}}, 2024.

\bibitem[Zhou et~al.(2024)Zhou, Pan, LeCun, and Pinto]{dinowm}
G.~Zhou, H.~Pan, Y.~LeCun, and L.~Pinto.
\newblock Dino-wm: World models on pre-trained visual features enable zero-shot planning.
\newblock \emph{arXiv preprint arXiv:2411.04983}, 2024.

\bibitem[Hansen et~al.(2024)Hansen, Su, and Wang]{tdmpc2}
N.~Hansen, H.~Su, and X.~Wang.
\newblock {TD-MPC2:} scalable, robust world models for continuous control.
\newblock In \emph{International Conference on Learning Representations, {ICLR}}, 2024.

\bibitem[Micheli et~al.(2023)Micheli, Alonso, and Fleuret]{iris}
V.~Micheli, E.~Alonso, and F.~Fleuret.
\newblock Transformers are sample-efficient world models.
\newblock In \emph{The Eleventh International Conference on Learning Representations, {ICLR}}, 2023.

\bibitem[Hafner et~al.(2020)Hafner, Lillicrap, Ba, and Norouzi]{dreamer}
D.~Hafner, T.~P. Lillicrap, J.~Ba, and M.~Norouzi.
\newblock Dream to control: Learning behaviors by latent imagination.
\newblock In \emph{International Conference on Learning Representations, {ICLR} 2020}, 2020.

\bibitem[Hafner et~al.(2021)Hafner, Lillicrap, Norouzi, and Ba]{dreamerv2}
D.~Hafner, T.~P. Lillicrap, M.~Norouzi, and J.~Ba.
\newblock Mastering atari with discrete world models.
\newblock In \emph{International Conference on Learning Representations, {ICLR}}, 2021.

\bibitem[Liu et~al.(2024)Liu, Zeng, Ren, Li, Zhang, Yang, Jiang, Li, Yang, Su, et~al.]{liu2024grounding}
S.~Liu, Z.~Zeng, T.~Ren, F.~Li, H.~Zhang, J.~Yang, Q.~Jiang, C.~Li, J.~Yang, H.~Su, et~al.
\newblock Grounding dino: Marrying dino with grounded pre-training for open-set object detection.
\newblock In \emph{European Conference on Computer Vision}, pages 38--55. Springer, 2024.

\bibitem[Oquab et~al.(2024)Oquab, Darcet, Moutakanni, Vo, Szafraniec, Khalidov, Fernandez, Haziza, Massa, El-Nouby, et~al.]{dinov2}
M.~Oquab, T.~Darcet, T.~Moutakanni, H.~Vo, M.~Szafraniec, V.~Khalidov, P.~Fernandez, D.~Haziza, F.~Massa, A.~El-Nouby, et~al.
\newblock Dinov2: Learning robust visual features without supervision.
\newblock \emph{Trans. Mach. Learn. Res.}, 2024, 2024.

\bibitem[Goel et~al.(2022)Goel, Bansal, Bhatia, Rossi, Vinay, and Grover]{goel2022cyclip}
S.~Goel, H.~Bansal, S.~Bhatia, R.~Rossi, V.~Vinay, and A.~Grover.
\newblock Cyclip: Cyclic contrastive language-image pretraining.
\newblock \emph{Advances in Neural Information Processing Systems}, 35:\penalty0 6704--6719, 2022.

\bibitem[Li et~al.(2022)Li, Li, Xiong, and Hoi]{li2022blip}
J.~Li, D.~Li, C.~Xiong, and S.~Hoi.
\newblock Blip: Bootstrapping language-image pre-training for unified vision-language understanding and generation.
\newblock In \emph{International conference on machine learning}, pages 12888--12900. PMLR, 2022.

\bibitem[Zhai et~al.(2023)Zhai, Mustafa, Kolesnikov, and Beyer]{siglip}
X.~Zhai, B.~Mustafa, A.~Kolesnikov, and L.~Beyer.
\newblock Sigmoid loss for language image pre-training.
\newblock In \emph{Proceedings of the IEEE/CVF International Conference on Computer Vision}, pages 11975--11986, 2023.

\bibitem[Ha and Schmidhuber(2018)]{ha2018world}
D.~Ha and J.~Schmidhuber.
\newblock World models.
\newblock \emph{arXiv preprint arXiv:1803.10122}, 2018.

\bibitem[Schmidhuber(1990{\natexlab{a}})]{schmidhuber1990line}
J.~Schmidhuber.
\newblock An on-line algorithm for dynamic reinforcement learning and planning in reactive environments.
\newblock In \emph{IJCNN international joint conference on neural networks}, pages 253--258. IEEE, 1990{\natexlab{a}}.

\bibitem[Schmidhuber(1990{\natexlab{b}})]{schmidhuber1990reinforcement}
J.~Schmidhuber.
\newblock Reinforcement learning in markovian and non-markovian environments.
\newblock \emph{Advances in neural information processing systems}, 3, 1990{\natexlab{b}}.

\bibitem[Racani{\`e}re et~al.(2017)Racani{\`e}re, Weber, Reichert, Buesing, Guez, Jimenez~Rezende, Puigdom{\`e}nech~Badia, Vinyals, Heess, Li, et~al.]{racaniere2017imagination}
S.~Racani{\`e}re, T.~Weber, D.~Reichert, L.~Buesing, A.~Guez, D.~Jimenez~Rezende, A.~Puigdom{\`e}nech~Badia, O.~Vinyals, N.~Heess, Y.~Li, et~al.
\newblock Imagination-augmented agents for deep reinforcement learning.
\newblock In \emph{Advances in neural information processing systems}, 2017.

\bibitem[Ding et~al.(2024)Ding, Zhang, Tian, and Zheng]{ding2024diffusion}
Z.~Ding, A.~Zhang, Y.~Tian, and Q.~Zheng.
\newblock Diffusion world model: Future modeling beyond step-by-step rollout for offline reinforcement learning.
\newblock \emph{arXiv preprint arXiv:2402.03570}, 2024.

\bibitem[Alonso et~al.(2024)Alonso, Jelley, Micheli, Kanervisto, Storkey, Pearce, and Fleuret]{alonso2024diffusion}
E.~Alonso, A.~Jelley, V.~Micheli, A.~Kanervisto, A.~J. Storkey, T.~Pearce, and F.~Fleuret.
\newblock Diffusion for world modeling: Visual details matter in atari.
\newblock \emph{Advances in Neural Information Processing Systems}, 37:\penalty0 58757--58791, 2024.

\bibitem[Hafner et~al.(2023)Hafner, Pasukonis, Ba, and Lillicrap]{dreamerv3}
D.~Hafner, J.~Pasukonis, J.~Ba, and T.~Lillicrap.
\newblock Mastering diverse domains through world models.
\newblock \emph{arXiv preprint arXiv:2301.04104}, 2023.

\bibitem[Schulman et~al.(2017)Schulman, Wolski, Dhariwal, Radford, and Klimov]{schulman2017proximal}
J.~Schulman, F.~Wolski, P.~Dhariwal, A.~Radford, and O.~Klimov.
\newblock Proximal policy optimization algorithms.
\newblock \emph{arXiv preprint arXiv:1707.06347}, 2017.

\bibitem[Haarnoja et~al.(2018)Haarnoja, Zhou, Abbeel, and Levine]{haarnoja2018soft}
T.~Haarnoja, A.~Zhou, P.~Abbeel, and S.~Levine.
\newblock Soft actor-critic: Off-policy maximum entropy deep reinforcement learning with a stochastic actor.
\newblock In \emph{International conference on machine learning}, pages 1861--1870. Pmlr, 2018.

\bibitem[Huang et~al.(2023)Huang, Qin, Liu, and Xu]{ddm}
Y.~Huang, Z.~Qin, X.~Liu, and K.~Xu.
\newblock Simultaneous image-to-zero and zero-to-noise: Diffusion models with analytical image attenuation.
\newblock \emph{arXiv preprint arXiv:2306.13720}, 2023.

\bibitem[Peebles and Xie(2023)]{peebles2023scalable}
W.~Peebles and S.~Xie.
\newblock Scalable diffusion models with transformers.
\newblock In \emph{Proceedings of the IEEE/CVF international conference on computer vision}, pages 4195--4205, 2023.

\bibitem[Liu et~al.(2024)Liu, Zhu, Gao, Feng, Liu, Zhu, and Stone]{libero}
B.~Liu, Y.~Zhu, C.~Gao, Y.~Feng, Q.~Liu, Y.~Zhu, and P.~Stone.
\newblock Libero: Benchmarking knowledge transfer for lifelong robot learning.
\newblock \emph{Advances in Neural Information Processing Systems}, 36, 2024.

\bibitem[Mees et~al.(2022)Mees, Hermann, Rosete-Beas, and Burgard]{mees2022calvin}
O.~Mees, L.~Hermann, E.~Rosete-Beas, and W.~Burgard.
\newblock Calvin: A benchmark for language-conditioned policy learning for long-horizon robot manipulation tasks.
\newblock \emph{IEEE Robotics and Automation Letters (RA-L)}, 7\penalty0 (3):\penalty0 7327--7334, 2022.

\bibitem[Tian et~al.(2025)Tian, Yang, Zeng, Wang, Lin, Dong, and Pang]{seer}
Y.~Tian, S.~Yang, J.~Zeng, P.~Wang, D.~Lin, H.~Dong, and J.~Pang.
\newblock Predictive inverse dynamics models are scalable learners for robotic manipulation.
\newblock In \emph{International Conference on Learning Representations}, 2025.

\bibitem[O’Neill et~al.(2024)O’Neill, Rehman, Maddukuri, Gupta, Padalkar, Lee, Pooley, Gupta, Mandlekar, Jain, et~al.]{o2024open}
A.~O’Neill, A.~Rehman, A.~Maddukuri, A.~Gupta, A.~Padalkar, A.~Lee, A.~Pooley, A.~Gupta, A.~Mandlekar, A.~Jain, et~al.
\newblock Open x-embodiment: Robotic learning datasets and rt-x models: Open x-embodiment collaboration 0.
\newblock In \emph{2024 IEEE International Conference on Robotics and Automation (ICRA)}, pages 6892--6903. IEEE, 2024.

\bibitem[Khazatsky et~al.(2024)Khazatsky, Pertsch, Nair, Balakrishna, Dasari, Karamcheti, Nasiriany, Srirama, Chen, Ellis, et~al.]{khazatsky2024droid}
A.~Khazatsky, K.~Pertsch, S.~Nair, A.~Balakrishna, S.~Dasari, S.~Karamcheti, S.~Nasiriany, M.~K. Srirama, L.~Y. Chen, K.~Ellis, et~al.
\newblock Droid: A large-scale in-the-wild robot manipulation dataset.
\newblock \emph{arXiv preprint arXiv:2403.12945}, 2024.

\bibitem[Tang et~al.(2024)Tang, Akinola, Xu, Wen, Handa, Wyk, Fox, Sukhatme, Ramos, and Narang]{tang2024automate}
B.~Tang, I.~Akinola, J.~Xu, B.~Wen, A.~Handa, K.~V. Wyk, D.~Fox, G.~S. Sukhatme, F.~Ramos, and Y.~S. Narang.
\newblock Automate: Specialist and generalist assembly policies over diverse geometries.
\newblock In D.~Kulic, G.~Venture, K.~E. Bekris, and E.~Coronado, editors, \emph{Robotics: Science and Systems}, 2024.

\bibitem[Kim et~al.(2024)Kim, Pertsch, Karamcheti, Xiao, Balakrishna, Nair, Rafailov, Foster, Sanketi, Vuong, Kollar, Burchfiel, Tedrake, Sadigh, Levine, Liang, and Finn]{openvla}
M.~J. Kim, K.~Pertsch, S.~Karamcheti, T.~Xiao, A.~Balakrishna, S.~Nair, R.~Rafailov, E.~P. Foster, P.~R. Sanketi, Q.~Vuong, T.~Kollar, B.~Burchfiel, R.~Tedrake, D.~Sadigh, S.~Levine, P.~Liang, and C.~Finn.
\newblock Openvla: An open-source vision-language-action model.
\newblock In \emph{Conference on Robot Learning}, pages 2679--2713, 2024.

\end{thebibliography}
\newpage
\section*{Appendix}
\appendix
\section{Architecture Details}\label{apd:1}
\begin{figure}[h]
    \centering
    \includegraphics[width=1\textwidth]{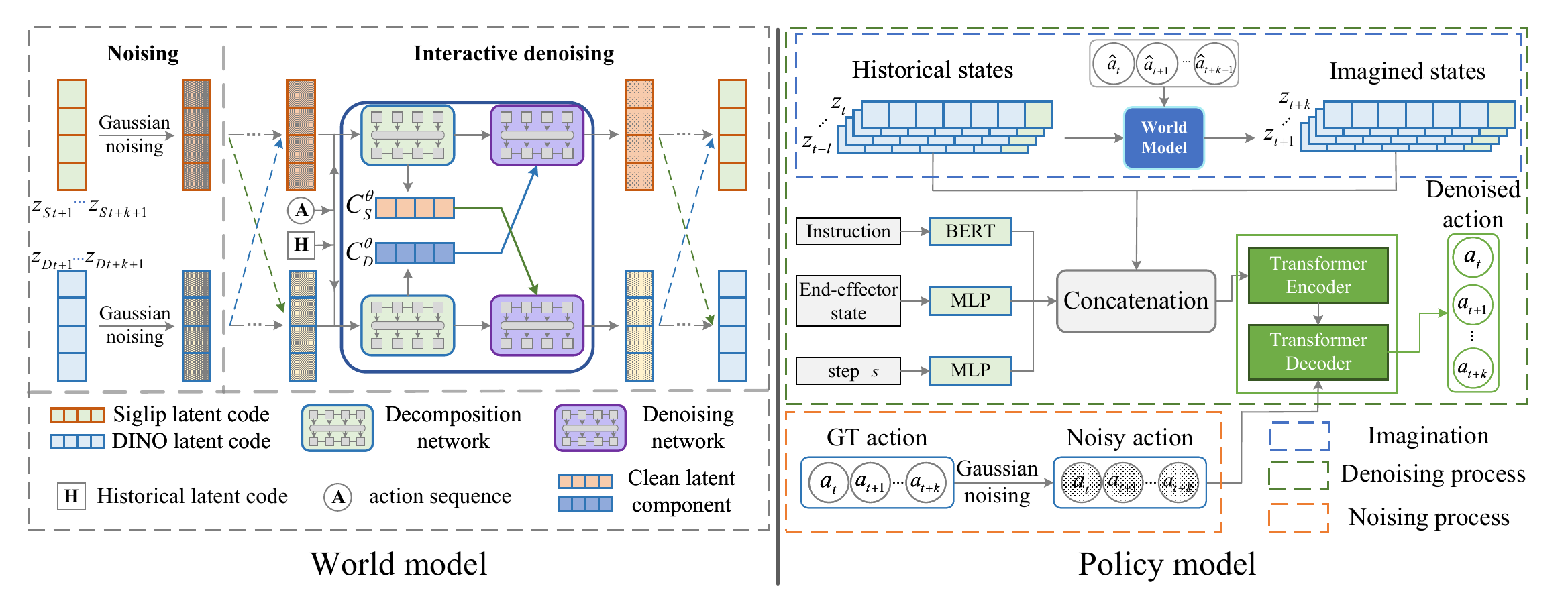}
    \vspace{-15pt}
    \caption{Detailed architecture of the world model and policy model.}
    \label{fig:arch}
\end{figure}

\textbf{World model}. As shown on the left of Fig.~\ref{fig:arch}, there are a noising process and a denoising process. We separately add the Gaussian noise to the DINO and Siglip latent codes in the noising process. For the denoising process, we adopt an 8-layer transformer with a hidden dimension of 384 as the decomposition network. For the denoising network, we utilize a cross-attention layer to let the two types of latent representations interact with each other.

We use the mean square error (MSE) loss to train the world model, and the training objective is formulated by:
\begin{equation}
\begin{aligned}
    \min\limits_{\boldsymbol{\theta}_1, \boldsymbol{\theta}_2, \boldsymbol{\theta}_3, \boldsymbol{\theta}_4} \mathbb{E}_{q(\mathbf{z}_{t+1:,0})} &\mathbb{E}_{q(\boldsymbol{\eta})} [\Vert {\mathbf{z}^{D}_{t+1:,\boldsymbol{\theta}}}-{\mathbf{z}^{D}_{t+1:,0}}\Vert^{2} + \Vert \boldsymbol{\eta}^{D}_{\boldsymbol{\theta}}-\boldsymbol{\eta}^{D}\Vert^{2} \\
    &\mathbb{E}_{q(\boldsymbol{\eta})} [\Vert {\mathbf{z}^{S}_{t+1:,\boldsymbol{\theta}}}-{\mathbf{z}^{S}_{t+1:,0}}\Vert^{2} + \Vert \boldsymbol{\eta}^{S}_{\boldsymbol{\theta}}-\boldsymbol{\eta}^{S}\Vert^{2} \\
    &+\Vert \boldsymbol{C}^{D}_{\boldsymbol{\theta}}-\boldsymbol{C}^{D}_{n}\Vert^{2}+\Vert \boldsymbol{C}^{S}_{\boldsymbol{\theta}}-\boldsymbol{C}^{S}_{n}\Vert^{2}]
\end{aligned}
\end{equation}
where $\boldsymbol{\theta}=[\boldsymbol{\theta}_1, \boldsymbol{\theta}_2, \boldsymbol{\theta}_3, \boldsymbol{\theta}_4]$ are the parameters of the diffusion world model.

\textbf{Policy network}. 
The policy model is depicted in the right of Fig.~\ref{fig:arch}. There are two processes in the diffusion policy: the noising and denoising processes. In the noising stage, we gradually add the Gaussian noise to the ground truth action until it is pure noise; while we leverage the transformer layers to learn the denoising process for recovering the action. 
We first tokenize the different modalities of inputs, which are fed into the transformer encoder and decoder to obtain the final output.
The transformer encoder and decoder contain 6 layers and 4 layers, and the hidden dimension is 256.
We utilize the MSE loss between the denoised action and the ground truth action to train the policy model.

\textbf{Implementation details.}
The length $l$ of historical videos is fixed at 4 frames, while the length of the imagination is ablated in Sec.~\ref{sec:ablation_study}. The world model is trained with a batch size of 4 on an NVIDIA 4090 GPU, which requires approximately three days. 
We utilize the same device to train the policy model with a batch size of 24, which takes approximately six hours. The raw image resolution is 128 $\times$ 128, and the patch size is 14$\times$14. The action space is seven-dimensional, representing the delta of translation, rotation, and aperture of the end-effector.

\section{Inference procedure}\label{apd:3}
In the inference stage, we first let the policy model generate an initial action, which is then fed into the world model to obtain the imagination guidance. Since the initial action may be suboptimal, the corresponding imagination guidance may not be accurate enough to provide effective guidance.
We adopt an iterative correction mechanism to refine the action based on progressively more accurate imagined states generated by the world model. This closed-loop interaction between policy refinement and world model prediction enables increasingly precise action optimization, leading to more robust and effective policy execution. The pseudocode is shown in Alg.~\ref{alg:inference}.
\begin{algorithm}[t]
  \caption{Pseudocode of inference procedure.}\label{alg:inference}
  \begin{multicols}{2}  
    \begin{algorithmic}[1]
      \State \textbf{Initialize} $t=$ 0, environment $env$, policy network $\pi_{\boldsymbol{\theta}}$, world model $p_{\boldsymbol{\theta}}$, $done=$ False, $max\_step=$ 600, instruction $\mathcal{T}$, refine iterations $m$.
        \State\textbf{Reset env}: $\mathbf{z}_0,\mathbf{e}_0=env.reset()$, $a_{0:k}=\boldsymbol{0}$;
        \While{not $done$ $|$ $t<max\_step$}
        \Statex
          \State\# iterative refinement loop:
          \For{$i \gets 1$ to $m$}
          \State\# get imagination:
          \State $\hat{\mathbf{z}}_{t+1:t+k+1}=p_{\boldsymbol{\theta}}(\mathbf{z}_{t-l:t},a_{t:t+k})$;
          \State \# get action:
          \State $a_{t:t+k} = \pi_{\boldsymbol{\theta}}(\mathbf{z}_{t-l:t}, \mathbf{e}_{t-l:t},$\\ 
          \quad\quad\quad\quad\quad$\hat{\mathbf{z}}_{t+1:t+k+1}, \mathcal{T})$;
          \EndFor
        
        \State \# environment step:
        \State  $\mathbf{z}_t, \mathbf{e}_t, done=env.step(a_t)$;
        \State $t=t+1$;
        
        \Statex
        
        \EndWhile
        
        
    \end{algorithmic}
  \end{multicols}
\end{algorithm}

\section{Experiment Setup}\label{apd:2}

\subsection{Benchmarks and Evaluation Metrics}\label{apd:dataset}
\textbf{LIBERO-LONG} is designed for language-conditioned long-horizon manipulation tasks and provides various suites with expert human demonstrations. We train the world model on the LIBERO-90 dataset, which includes demonstrations for 90 short-horizon tasks. Note that the world model does not access the tasks of policy learning. Following \citet{seer}, we report the average success rate (Avg.SR) of the top-3 checkpoints over 20 rollouts for the evaluation. Here is the list of tasks: \\
$\bullet$ Task1: Turn on the stove and put the moka pot on it. \\
$\bullet$ Task2: Put the black bowl in the bottom drawer of the cabinet and close it. \\
$\bullet$ Task3: Put the yellow and white mug in the microwave and close it. \\
$\bullet$ Task4: Put both moka pots on the stove. \\
$\bullet$ Task5: Put both the alphabet soup and the cream cheese box in the basket. \\
$\bullet$ Task6: Put both the alphabet soup and the tomato sauce in the basket. \\
$\bullet$ Task7: Put both the cream cheese box and the butter in the basket. \\
$\bullet$ Task8: Put the white mug on the left plate and put the yellow and white mug on the right plate. \\
$\bullet$ Task9: Put the white mug on the plate and put the chocolate pudding to the right of the plate. \\
$\bullet$ Task10: Pick up the book and place it in the back compartment of the caddy. \\

\textbf{CALVIN D-D} is a benchmark focusing on language-conditioned visual robot manipulation, which contains 34 tasks varying in object and scene visual appearance. To make the training data of the world model different from that of the policy model, we selected half of the tasks, including `push', `move', `open', and `place', to train the world model. The remaining tasks, such as `lift', `rotate', and `stack', are used to evaluate policy learning. Following \citet{seer}, we report the average success rate and the average length (Avg.Len) of completed sequences.

\subsection{Baseline methods}
\begin{itemize}
    \item  \textbf{DreamerV3} \citep{dreamerv3} and \textbf{TDMPC2} \cite{tdmpc2} employ learned world models for reinforcement learning and planning. While these methods also operate in latent space, their representations are optimized for image reconstruction or reward prediction, neglecting critical geometric and semantic features essential for manipulation tasks. In contrast, our approach harnesses visual foundation models to derive latent representations that explicitly capture both geometric and semantic dynamics, enabling more effective learning for manipulation.
    In particular, since DreamerV3 and TDMPC2 require dense rewards for world model training, we follow \citet{tang2024automate} to design appropriate dense rewards for manipulation tasks. Additionally, given that these reinforcement learning-based methods struggle with manipulation tasks, we initialize their policies using behavior cloning.
    \item  \textbf{ATM} \cite{atm}. ATM employs a language-conditioned trajectory prediction model that provides point-based trajectories as policy inputs. However, this approach faces two key limitations: (1) the language grounding inherently restricts generalization to unseen tasks, and (2) the sparse point representation fails to capture critical observation details. Differently, our world model offers two distinct advantages: (1) it learns from task-agnostic video clips, enabling broader task generalization; (2) it operates in a rich latent space preserving geometric and semantic information.
    \item \textbf{Seer} \cite{seer}. Seer represents the previous state-of-the-art method on both the LIBERO-LONG and CALVIN benchmarks. The approach introduces an inverse dynamics model that jointly predicts actions and subsequent frame images. However, this implicit dynamics model cannot directly provide future state guidance for policy learning. In contrast, our world model allows the policy model to obtain future states through imagination, enabling iterative refinement of the action prediction.
\end{itemize}

\section{More Quantitative Comparisons}\label{apd:full_compare}

\paragraph{Comparison on LIBERO-LONG with full data.} While the main paper demonstrates our method's superiority under limited action-labeled data conditions, we additionally report full-data performance to establish the upper bound of our approach. As shown in Tab.~\ref{tab:all_data}, our method achieves the best average success rate of 90.7\%, representing a 3\% absolute improvement over the previous state-of-the-art.
\begin{table*}[t]
    \centering
    \caption{Performance comparisons on LIBERO-LONG benchmark with all training data.}
    \label{tab:all_data}
    \renewcommand{\arraystretch}{1}
    \resizebox{1\textwidth}{!}{
    \begin{tabular}{cccccc}
\toprule
Method                                                                   &
Avg.SR$\uparrow$  & \begin{tabular}[c]{@{}c@{}}Put bowl in\\ drawer and\\ close it\end{tabular}   & \begin{tabular}[c]{@{}c@{}}Put mug in\\ microwave and\\ close it\end{tabular}         & \begin{tabular}[c]{@{}c@{}}Put both pots\\ on stove\end{tabular}           & \begin{tabular}[c]{@{}c@{}}Put soup\\ and box\\ in basket\end{tabular} \\ \hline\hline
DreamerV3                                                &   39.3                                                             &           58.3                                                             &           25.0                                                                    &       30.0                                                          &    35.0                                                               \\
TDMPC2                                                 &     46.8                                                            &           61.7                                                             &           33.3                                                                       &                 40.0                                                      &    36.7                                                               \\
OpenVLA \citep{openvla}                                                 &     54.0                                                            &           45.0                                                             &           55.0                                                                       &                 20.0                                                      &    35.0                                                               \\
ATM                                                 &      59.6                                                         &           83.3                                                             &           45.0                                                                       &                 43.3                                                      &    60.0                                                               \\
Seer                                                                     &                                                 87.7                       &             \textbf{100}                                                              &               71.7                                                                    &             61.7                                                        &                       91.7                                         \\
\rowcolor{gray!20}
Ours                                                                     &        \textbf{90.7}                                                          &                    93.3                                                           &                     \textbf{95.0}                                                                  &                         \textbf{91.7}                                           &       \textbf{95.0}                                                                 \\ \hline\hline
\begin{tabular}[c]{@{}c@{}}Put soup\\ and sauce\\ in basket\end{tabular} & \begin{tabular}[c]{@{}c@{}}Put box\\ and buffer\\ in basket\end{tabular} & \begin{tabular}[c]{@{}c@{}}Put mugs on\\ left and\\ right plates\end{tabular} & \begin{tabular}[c]{@{}c@{}}Put mug on\\ plate and put\\ pudding to right\end{tabular} & \begin{tabular}[c]{@{}c@{}}Pick book\\ and place\\ it in back\end{tabular} & \begin{tabular}[c]{@{}c@{}}Turn on\\ stove and\\ put pot\end{tabular}                                                                     \\ \hline\hline
43.3                                                         &          33.3                                                          &       45.0                                                                  &                36.7                                                                  &             30.0                &              56.7    \\
45.0                                                    &           46.7                                                          &       58.3                                                                  &                45.0                                                                  &             40.0                &              61.7    \\
45.0                                                         &          \textbf{95.0}                                                          &       40.0                                                                  &                60.0                                                                  &             80.0                &              65.0    \\
53.3                                                         &           55.0                                                          &       63.3                                                                  &                60.0                                                                  &             45.0                &          88.3   \\
88.3                                                                     &           90.0                                                         &                        \textbf{91.7}                                                 &                                           85.0                              &                                    \textbf{93.3}                                        &                          98.3                                       \\
\rowcolor{gray!20}
\textbf{88.3}                                                                     &            90.0                                                              &                    85.0                                                           &                            \textbf{85.0}                                                           &                          88.3                                                  &          95.0                                                              \\ \bottomrule
\end{tabular}
    }
\end{table*}

\textbf{Cross-scene generalization with world model}.
To assess the generalization capacity of our world model, we directly transfer the model trained on the LIBERO benchmark to policy learning on the CALVIN benchmark, despite their distinct scene configurations. As demonstrated in the fifth row of Tab.~\ref{tab:calvin2}, our cross-scene implementation achieves an average sequence length (Avg.Len.) of 3.05, surpassing both vanilla behavior cloning (BC) by 0.61 and ATM by 0.07. These results substantiate our world model's ability to generalize across different environmental scenarios.

\begin{table*}[t]
    \centering
    \caption{Comparisons against state-of-the-art methods on CALVIN D-D benchmark.}
    \label{tab:calvin2}
    \renewcommand{\arraystretch}{1}
    \setlength{\tabcolsep}{8pt}
\begin{tabular}{l|cccccc}
\toprule
\multirow{2}{*}{Method} & \multicolumn{6}{c}{Task completed in a row}       \\ \cline{2-7} 
                        & 1 & 2 & 3 & 4 & \multicolumn{1}{c|}{5} & Avg.Len.$\uparrow$ \\ \hline\hline
Vanilla BC                     & 81.4  & 60.8  & 45.9  & 32.1  & \multicolumn{1}{c|}{24.2}  &  2.44    \\
DreamerV3  \citep{dreamerv3}                    & 82.0  & 63.1  & 46.6  & 34.1  & \multicolumn{1}{c|}{25.1}  &  2.51    \\
ATM   \citep{atm}                   & 83.3  & 70.9  & 58.5  & 46.3  & \multicolumn{1}{c|}{39.1}  &  2.98    \\
Seer  \citep{seer}                   & 92.2  & 82.6 &  71.5 & 60.6  & \multicolumn{1}{c|}{53.5}  &  3.60    \\
\rowcolor{gray!20}
Ours (cross-scene)                 & 84.1 & 72.2  & 60.3 & 48.5  & \multicolumn{1}{c|}{40.3}  &  3.05    \\ 
\rowcolor{gray!20}
Ours                    & \textbf{92.7} & \textbf{83.1}  &  \textbf{72.1} & \textbf{61.2}  & \multicolumn{1}{c|}{\textbf{54.1}}  &  \textbf{3.63}    \\ 
\bottomrule
\end{tabular}
\end{table*}

\section{Analysis of Iterative Refinement}\label{ape:convergence}
The proposed latent diffusion-based world model allows the policy model to refine itself iteratively, and the results in the main paper show it can significantly surpass the vanilla BC by 27.9\%. Here we conduct a series of experiments to analyze the effect of iterative refinement.

\textbf{Why does the iterative refinement improve the success rate?}
Generally speaking, the output of the diffusion policy is not a deterministic value but a stochastic distribution. 
We claim that the iterative refinement can gradually reduce the entropy of the predicted distribution, thereby aligning predictions more closely with ground truth at each execution step and consequently improving task success rates.
To validate this claim, we perform 10 repeated policy evaluations under different iteration settings. We leverage PCA to reduce the action dimension and Kernel Density Estimation to plot 4 distributions. As shown in Fig.~\ref{fig:1}, as the number of iterations increases, the entropy of the predicted distribution gradually reduces.

\begin{table}[hb]
        \begin{minipage}{0.5\textwidth}
        \caption{Performances of iterative refinement trained with different data scales. The results are obtained under 10 repeated evaluations.} 
        \label{tab:1}
        \centering
        \resizebox{\textwidth}{!}
        {
        \renewcommand{\arraystretch}{1.5}
        \setlength{\tabcolsep}{2pt}
        \begin{tabular}{lccc}
        \toprule
        Data scale & 10 demo & 20 demo & 30 demo \\ \hline\hline
        1-iter    & 59.9$\pm$7.23 & 75.1$\pm$6.65& 80.4$\pm$4.68                 \\
        2-iter    & 68.1$\pm$4.38 & 77.3$\pm$4.63& 81.5$\pm$3.82    \\
        3-iter    &68.3$\pm$3.25 & 77.5$\pm$4.01& 81.5$\pm$2.63    \\ 
        4-iter   & 68.6$\pm$2.51 & 77.5$\pm$3.12 &  81.7$\pm$1.93   \\ 
        \bottomrule
        \end{tabular}
        }
        \end{minipage}\quad\ \
        \begin{minipage}{0.4\textwidth}
        \centering
        \includegraphics[width=1.0\linewidth]{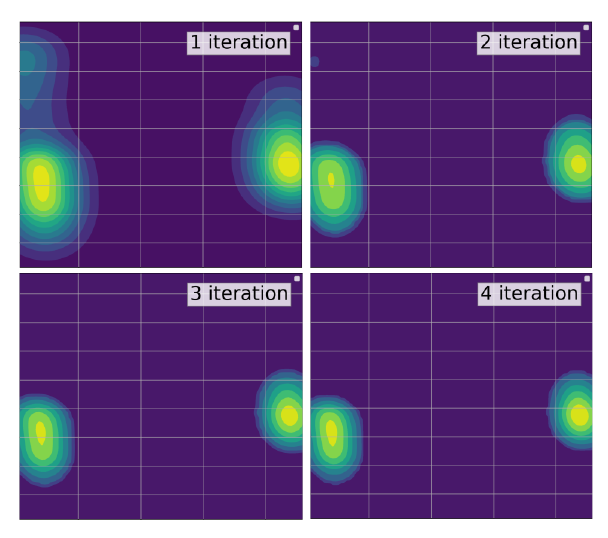}
        \captionof{figure}{Predicted action distribution. 
        }
        \label{fig:1}
        \end{minipage}
    \end{table}

\textbf{Convergence of iterative refinement.}
We conduct additional experiments with varying numbers of training demonstrations to further verify convergence properties. As shown in Tab.~\ref{tab:1}, our method consistently achieves convergence within 2 refinement iterations across all experimental settings. Furthermore, the observed variance continuously decreases with increasing iterations, demonstrating that iterative refinement enhances performance through increasingly precise (low-entropy) predictions.

\section{Ablation Studies}\label{apd:ablation}
\textbf{Effect of the world model}. We employ an imagination-guided policy based on behavior cloning (BC). To intuitively demonstrate the effect of world model guidance, we compare our method with a vanilla behavior cloning policy that solely relies on historical observations as inputs. Besides, we implement a comparison model that utilizes a copy of the current observation as the future imagination, and we call it the copy model. Our world model, which serves as a simulated environment, can provide future imagined states as efficient guidance.
As reported in Tab.~\ref{tab:2}, the copy model can not bring improvement and is even slightly worse than vanilla behavior cloning. In contrast, our approach improves the vanilla behavior cloning method by 27.9\%, highlighting the significant potential of imagination-guided policy learning. 


\begin{table}[t]
    \centering
    \caption{Ablation on the effect of the world model.}
    \label{tab:2}
    \renewcommand{\arraystretch}{1.2}
    \setlength{\tabcolsep}{4pt}
    \resizebox{1\linewidth}{!}{
    \begin{tabular}{lccccccccccc}
    \toprule
    Method  & Task1 & Task2 & Task3 & Task4 & Task5 &Task6 & Task7& Task8& Task9&Task10 & Avg.SR \\\hline\hline
    BC     &  66.7 &  63.3 &  31.7  &  40.0   & 35.0 & 41.7 & 35.0  & 43.3 & 25.0 & 26.7 & 40.8  \\
    Copy model     &  68.3 &  40.0 &  51.7  &  36.7   & 41.7 & 26.7 & 50.0  & 20.0 & 30.0 & 36.7 & 40.2\\
    \rowcolor{gray!20}
    Ours  &  \textbf{88.3} &  \textbf{68.3} &  \textbf{63.3}  &  \textbf{45.0}   & \textbf{83.3} & \textbf{65.0} & \textbf{78.3}  & \textbf{63.3} & \textbf{60.0} & \textbf{71.7} & \textbf{68.7} \\
    \bottomrule
    \end{tabular}
    }
\end{table}

\textbf{Latent diffusion vs. image pixel diffusion}. Unisim \cite{unisim} proposes a pixel diffusion world model; however, it is not open-source. To ensure a fair comparison, we also employ our diffusion world model in the pixel space, where we retain the network architecture but remove the latent representation extraction stage. Tab.~\ref{tab:3} presents a detailed performance comparison. Our method learns the diffusion process in a compact and more generalizable latent space, leveraging the features of pretrained foundation models, which results in robust dynamic predictions for unseen tasks. In contrast, the pixel diffusion world model exhibits inferior performance, with a 14.7\% drop in the average success rate. 


\begin{table}[t]
    \centering
    \caption{Ablation on different diffusion spaces.}
    \label{tab:3}
    \renewcommand{\arraystretch}{1.5}
    \setlength{\tabcolsep}{4pt}
    \resizebox{1\linewidth}{!}{
    \begin{tabular}{lccccccccccc}
    \toprule
    Method  & Task1 & Task2 & Task3 & Task4 & Task5  &Task6 & Task7& Task8& Task9&Task10 & Avg.SR \\\hline\hline
    Pixel diffusion    &  83.3 &  \textbf{78.3} &  50.0  &  35.0   & 61.7 & 50.0 & 45.0  & \textbf{65.0} & 36.7 & 35.0 & 54.0 \\
    \rowcolor{gray!20}
    Ours  &  \textbf{88.3} &  68.3 &  \textbf{63.3}  &  \textbf{45.0}   & \textbf{83.3} & \textbf{65.0} & \textbf{78.3}  & 63.3 & \textbf{60.0} & \textbf{71.7} & \textbf{68.7} \\\bottomrule
    \end{tabular}
    }
\end{table}

\begin{table}[ht]
    \centering
    \caption{Ablation on imagination frames and denoising steps. The results are obtained with 1-iter refinement.}
    \label{tab:5}
    \renewcommand{\arraystretch}{1.3}
    \setlength{\tabcolsep}{5pt}
    \begin{tabular}{lccccc|lccccc}
    \toprule
    Frames & 1 & 2 & 4 & 6 & 8 & Denoising step & 1 & 2 & 5 & 10 & 30 \\
    \hline
    Avg.SR & 44.9  & 49.6  &  56.3 &  60.6  & 60.0 & Avg.SR         & 60.0 & 60.6  &  60.5 & 60.3   &  60.6  \\ \bottomrule
    \end{tabular}
\end{table}

\begin{table}[ht]
    \centering
    \caption{Ablation on different robotic states as the inputs of the policy network. The results are obtained with 1-iter refinement.}
    \label{tab:6}
    \renewcommand{\arraystretch}{1.5}
    \setlength{\tabcolsep}{4pt}
    \resizebox{1\linewidth}{!}{
    \begin{tabular}{lccccccccccc}
    \toprule
    Method  & Task1 & Task2 & Task3 & Task4 & Task5  &Task6 & Task7& Task8& Task9&Task10 & Avg.SR \\\hline\hline
    EE states    & 75.0 &  61.7 &  50.0  &  43.3   & 55.0 & 51.7 & 50.0  & 65.0 & 35.0 & 30.0 & 51.7  \\
    \rowcolor{gray!20}
    joint states   &90.0 &  83.3 &  50.0  &  45.0   & 66.7 & 55.0 & 55.0  & 71.7 & 50.0 & 40.0 & 60.7   \\\bottomrule
    \end{tabular}
    }
\end{table}

\textbf{Effect of the imagined future frames}. The imagination-guided policy network takes the future latent states generated by the proposed world model as inputs. We conduct a series of experiments to investigate the impact of varying numbers of imagined latent state frames, and the results are presented in Tab.~\ref{tab:5}. As the number of frames increases, the average success rate gradually improves, peaking at 6 frames. We hypothesize that excessively long future imagination may become unreliable due to compounding errors.

\textbf{Effect of different denoising steps}. For the policy network, we utilize ADM \cite{ddm} as our diffusion algorithm, which introduces an analytical attenuation process to speed up the denoising process. We report the results of different denoising steps in Tab.~\ref{tab:5}. Benefiting from the fast diffusion algorithm, our policy network can achieve a convergent performance with only 2 denoising steps.

\textbf{Effect of different robotic states}. There are two common options for the robotic states as inputs to the policy network: joint states and end-effector states (EE states). As shown in Tab.~\ref{tab:6}, the use of joint states outperforms EE states by 9.0\%, suggesting that joint states are more suitable for solving manipulation tasks.


\section{More Details of Real-world experiments}\label{apd:real}
\textbf{Task definition.} We define 7 manipulation tasks in the real scene, in order: `stack the middle bowl on the back bowl', `put the bowl inside the middle drawer of the cabinet and close it', `open the top drawer and put the bowl inside', `pick up the milk and place it in the basket', `pick up the cans and place it in the basket', `pick up the orange juice and place it in the basket', `pick up the ketchup and place it in the basket'.

\begin{figure}
    \centering\vspace{-5pt}
    \includegraphics[width=1\textwidth]{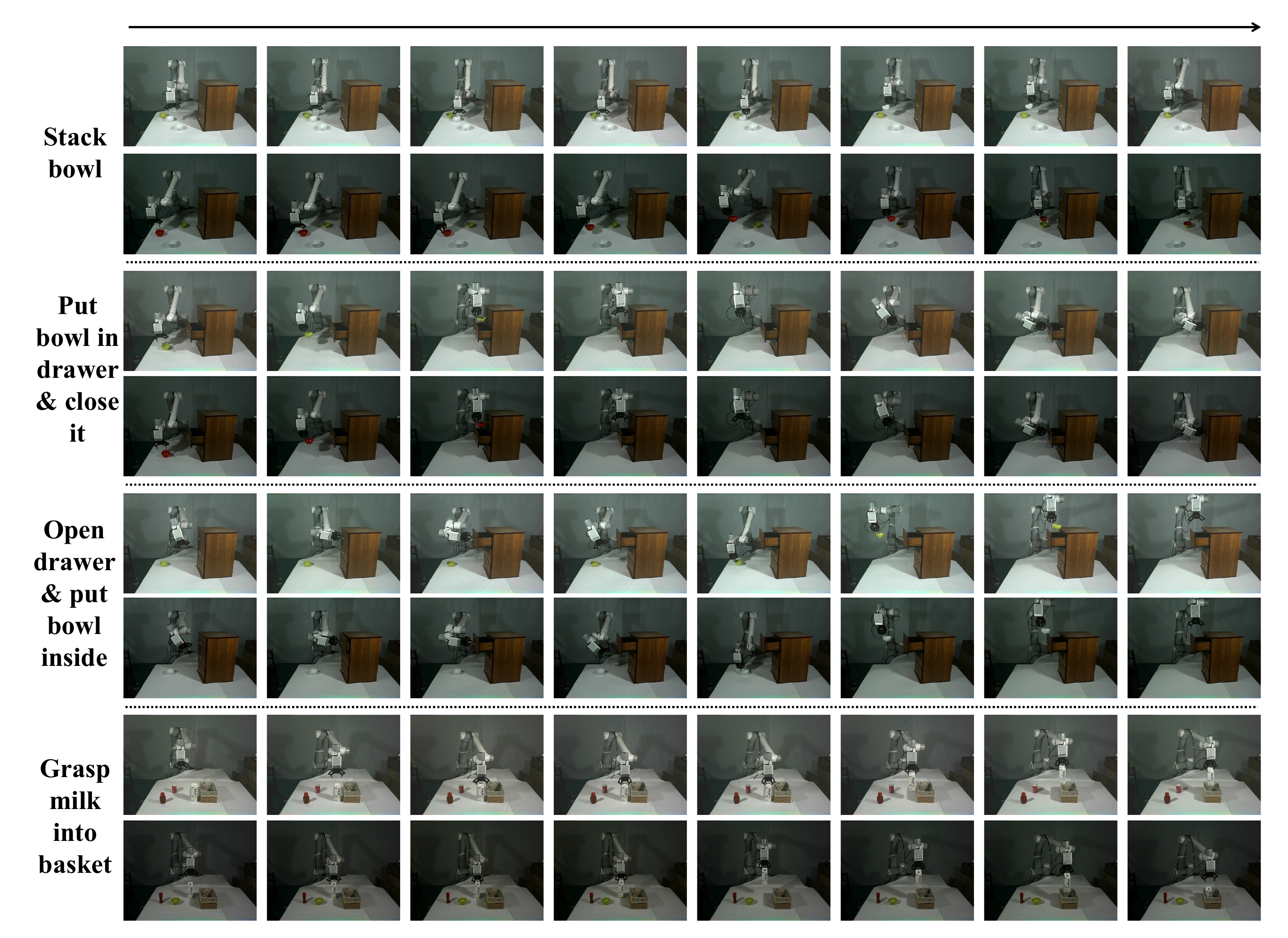}\vspace{-5pt}
    \caption{More visual results of real-world manipulation.}
    \label{fig:real_vis}
\end{figure}

\begin{figure}
    \centering\vspace{-5pt}
    \includegraphics[width=1\textwidth]{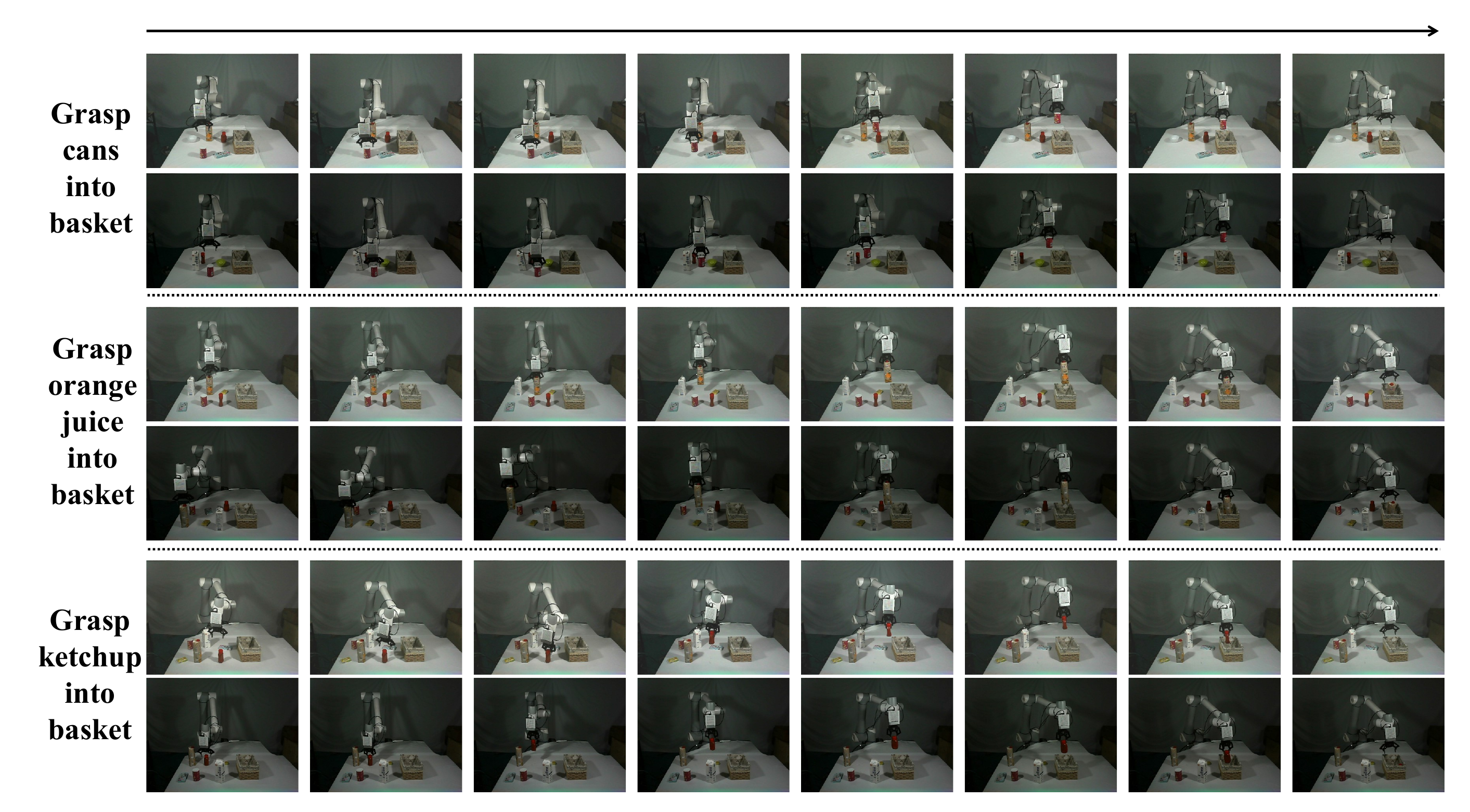}\vspace{-5pt}
    \caption{More visual results of real-world manipulation.}
    \label{fig:real_vis2}
\end{figure}

\textbf{More visual results of manipulation trajectory.}
We depict more visual trajectories of the real-world manipulation results in Fig.~\ref{fig:real_vis} and Fig.~\ref{fig:real_vis2}. Benefiting from the powerful dynamic modeling ability of the world model, our method can deal with complex lighting changes.
\newpage

\end{document}